\pdfoutput=1

\documentclass[11pt]{article}

\usepackage[]{ACL2024}

\usepackage{times}
\usepackage{latexsym}

\usepackage[T1]{fontenc}

\usepackage[utf8]{inputenc}

\usepackage{microtype}

\usepackage{inconsolata}

\usepackage{booktabs}
\usepackage{multirow}
\usepackage{multicol}
\usepackage{makecell}
\usepackage{enumitem}
\usepackage{balance}
\usepackage{graphicx}
\usepackage{subfigure}
\usepackage{amsmath}
\usepackage{amsthm}
\usepackage{amsfonts}
\usepackage{hyperref}
\usepackage{arydshln}

\usepackage[breakable,skins]{tcolorbox}
\usepackage{alltt}
\usepackage{xcolor}
\newtcolorbox{AIBox}[2][]{aibox,title=#2,#1}

\tcbset{
  aibox/.style={
    width=0.95\textwidth,
    top=5pt,
    colback=black!05,
    colframe=black!20,
    colbacktitle=black!50,
    enhanced,
    center,
    attach boxed title to top left={yshift=-0.1in,xshift=0.1in},
    boxed title style={boxrule=0pt,colframe=white,},
  }
}

\title{\textit{Instruct Once, Chat Consistently in Multiple Rounds}: An Efficient \\ Tuning Framework for Dialogue}

\author{Jian Wang\textsuperscript{\rm 1},
    Chak Tou Leong\textsuperscript{\rm 1},
    Jiashuo Wang\textsuperscript{\rm 1},
    Dongding Lin\textsuperscript{\rm 1}, \\
    \textbf{Wenjie Li}\textsuperscript{\rm 1},
    \textbf{Xiao-Yong Wei}\textsuperscript{\rm 2,1} \\
    \textsuperscript{\rm 1}Department of Computing, The Hong Kong Polytechnic University \\
    \textsuperscript{\rm 2}School of Computer Science, Sichuan University \\
    \texttt{\{jian-dylan.wang,
   chak-tou.leong,
   dongding88.lin\}@connect.polyu.hk} \\
   \texttt{\{csjwang, cswjli\}@comp.polyu.edu.hk} ~~ \texttt{cswei@scu.edu.cn}
}

\begin{document}
\maketitle
\begin{abstract}
Tuning language models for dialogue generation has been a prevalent paradigm for building capable dialogue agents. Yet, traditional tuning narrowly views dialogue generation as resembling other language generation tasks, ignoring the role disparities between two speakers and the multi-round interactive process that dialogues ought to be. Such a manner often leads to unsatisfactory chat consistency for the built agent. In this work, we emphasize the interactive, communicative nature of dialogue and argue that it is more feasible to model the speaker roles of agent and user separately, enabling the agent to adhere to its role consistently. With this in mind, we propose an efficient \textbf{M}ulti-round \textbf{I}nteractive \textbf{Di}alogue Tuning (\textsc{Midi}-Tuning) framework\footnote{Our code and data are available at \url{https://github.com/iwangjian/Midi-Tuning}.}. It models the agent and user individually with two adapters built upon large language models. The adapters make use of respective utterances round by round in alternating order and they are tuned via a round-level memory caching mechanism. Extensive experiments demonstrate that, our framework performs superior to traditional fine-tuning and harbors the tremendous potential for improving dialogue consistency.

\end{abstract}

\section{Introduction}

\begin{figure}[t!]
\centering
\subfigure[One-dialogue-$n$-sample: split a multi-round dialogue into multiple single-round yet non-independent samples.]{
    \begin{minipage}[b]{0.93\linewidth}
	    \includegraphics[width=1\linewidth]{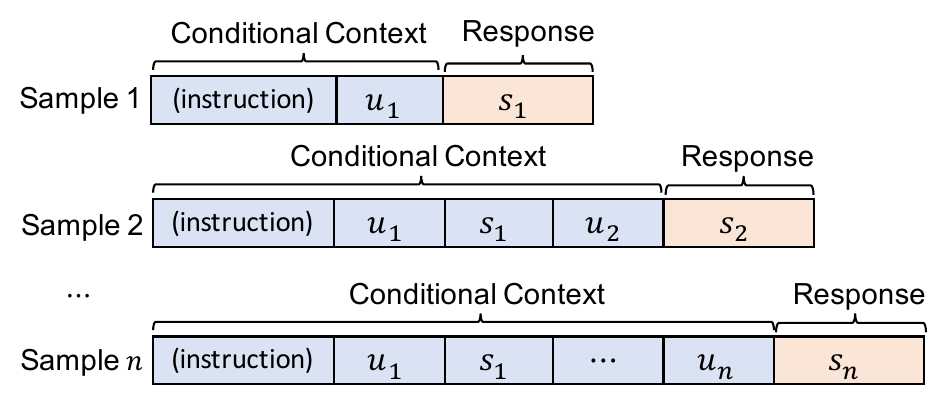}
    \end{minipage}
    \label{fig:paradigm_a}
}
\subfigure[One-dialogue-one-sample: utilize a multi-round dialogue sample at once based on causal masks.]{
    \begin{minipage}[b]{0.975\linewidth}
	    \includegraphics[width=1\linewidth]{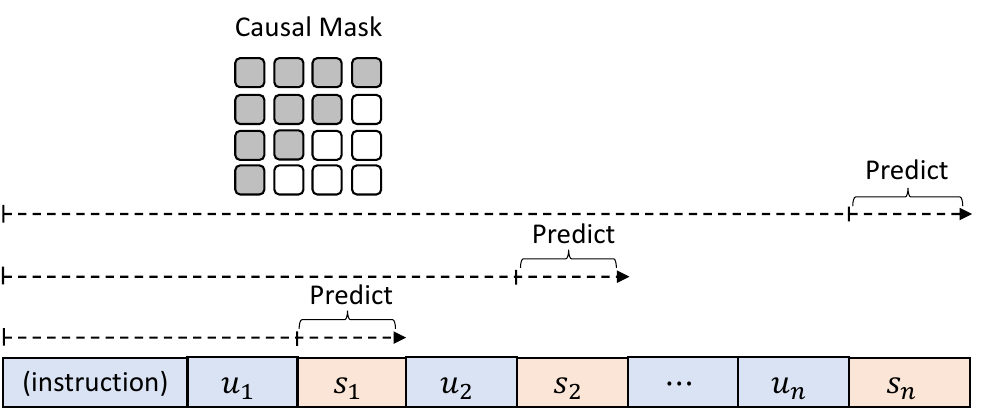}
    \end{minipage}
    \label{fig:paradigm_b}
}
\subfigure[Our \textsc{Midi}-Tuning: utilize dialogue utterances round by round with an interactive process.]{
    \begin{minipage}[b]{0.92\linewidth}
	    \includegraphics[width=1\linewidth]{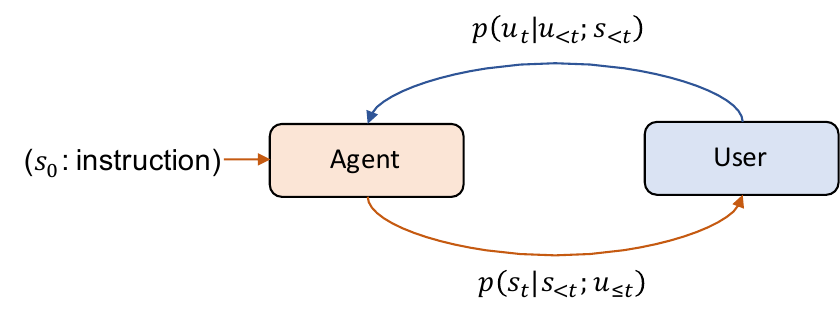}
    \end{minipage}
    \label{fig:paradigm_c}
}
\caption{Comparison of different tuning manners (including data usage) for dialogue generation.}
\label{fig:paradigm_comparison}
\end{figure}

Building human-like intelligent dialogue agents is a long-standing ambition for the research community of dialogue systems. Recently, we have witnessed a substantial revolution in advanced conversational agents such as ChatGPT \cite{openai2022chatgpt} and GPT-4 \cite{openai2023gpt4}, which are fundamentally built upon large language models (LLMs) \cite{brown2020language,bommasani2021opportunities}. Similar efforts have also been made by academia and open-source communities, leading to a variety of notable chat language models, such as Vicuna \cite{vicuna2023}, Koala \cite{koala_blogpost_2023}, and \textsc{Llama 2}-Chat \cite{touvron2023llama2}. These chat language models can be attained by instruction fine-tuning on downstream dialogue data, demonstrating promising performance in generating natural and comprehensive responses. Tuning LLMs for dialogue generation has been the de-facto mainstream practice towards creating capable dialogue agents.

Traditional dialogue tuning narrowly views dialogue generation as resembling other language generation tasks without distinction. It performs in either \textit{one-dialogue-$n$-sample} (see Figure \ref{fig:paradigm_a}) or \textit{one-dialogue-one-sample} (see Figure \ref{fig:paradigm_b}) manner. The former transforms dialogue model training into general language generation via splitting each multi-round dialogue into multiple single-round samples, yet results in non-independent distributions among those samples. The latter enhances training efficiency by utilizing each multi-round dialogue at once, which computes the prediction loss for the agent's responses through causal masks, such as Vicuna \cite{vicuna2023} and UltraLLaMA \cite{ding-etal-2023-enhancing}. However, these methods simply concatenate utterances from two speakers (e.g., user and agent) together (and instructions for the agent, if any)  and mix their content in the same language model space, ignoring the role disparities between two speakers and the multi-round interactive process that dialogues ought to be. Such tuning methods inevitably hinder a built dialogue agent from maintaining the chat \textit{consistency} \cite{touvron2023llama2,lu2023memochat}, requiring that the agent always adhere to its role even with the dialogue rounds moving forward. It remains urgent to solve for many consistency-demanding scenarios.

One of the primary challenges for improving dialogue consistency lies in the disparity modeling of the two speaker roles. It is because the inconsistency issue in real-world human communication \cite{wu-etal-2021-alternating,bao-etal-2022-learning,takmaz-etal-2023-speaking} is often caused by various types of speaker disparities, such as background knowledge, cognitive level, personalities, and goals. We emphasize that it is more feasible to \textit{model the roles of agent and user separately} (see Figure \ref{fig:paradigm_c}), such that the agent and user models can \textit{consistently adhere to their respective roles and interact with each other round by round}, similar to humans.

When tuning LLMs for conversation, we have a similar motivation towards consistent dialogue generation. We propose a general, simple, and effective framework, namely \textbf{M}ulti-round \textbf{I}nteractive \textbf{Di}alogue Tuning (\textsc{Midi}-Tuning). It employs two language model adapters (e.g., LoRA \cite{hu2022lora}) built upon LLMs, to represent the agent and user, respectively. The two adapters are tuned by utilizing respective utterances round by round in alternating order, with each adapter learning to distinguish language model distribution about its role. However, such separate modeling is non-trivial in tracking the complete dialogue context. Considering that the foundation architecture of mainstream LLMs is Transformer \cite{vaswani2017attention}, we propose a \textit{round-level memory caching} mechanism to address it efficiently, which reuses previous-round cached keys and values as ongoing context when processing present-round utterance. 

In summary, our main contributions are as follows: (1) To the best of our knowledge, this is the first work investigating how a new way of tuning could affect dialogue consistency in the era of LLMs. (2) We propose \textsc{Midi}-Tuning, a general, simple, and efficient framework to tune LLMs for dialogue generation, which can be applied in broad downstream dialogue scenarios. (3) Extensive experiments demonstrate that \textsc{Midi}-Tuning outperforms traditional fine-tuning over various LLMs, especially in maintaining consistency for multi-round dialogues.

\section{Related Work}

\begin{figure*}[th!]
\centering
\includegraphics[width=0.98\textwidth]{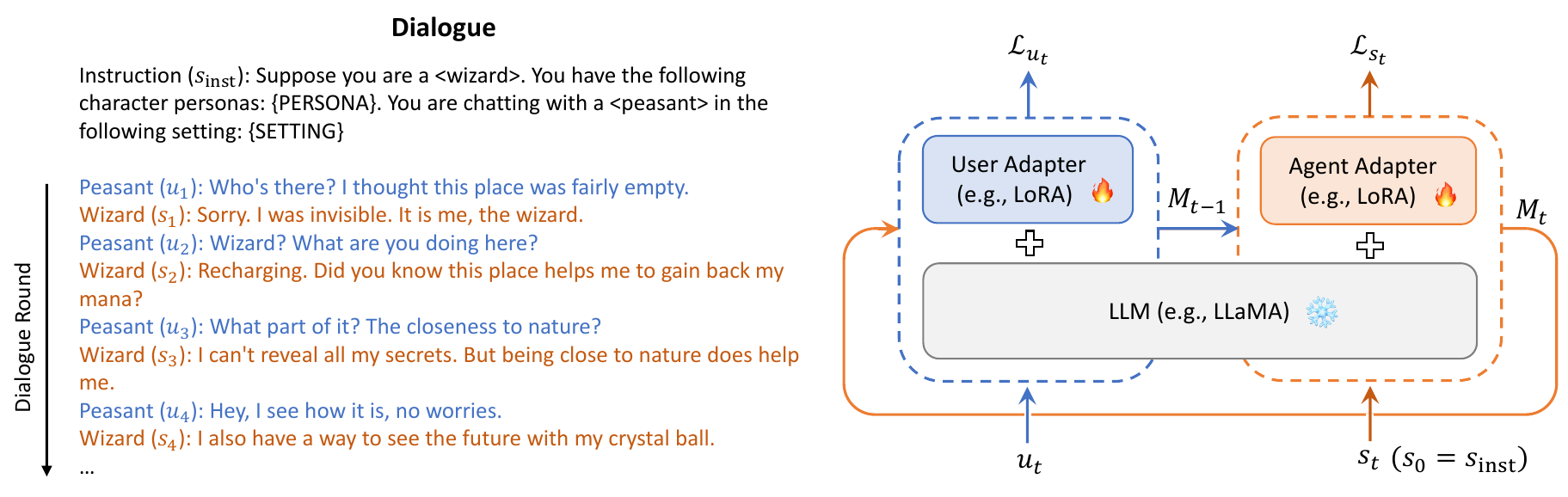}
\caption{Overview of the proposed Multi-round Interactive Dialogue Tuning (\textsc{Midi}-Tuning) framework.}
\label{fig:framework}
\end{figure*}

\begin{figure}[th!]
\centering
\includegraphics[width=0.82\linewidth]{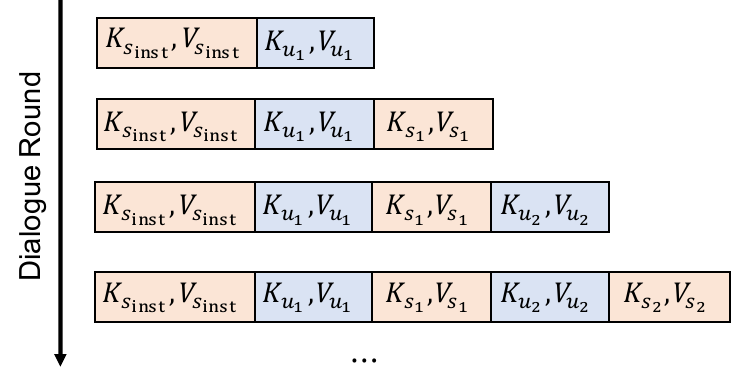}
\caption{Overview of the round-level memory caching.}
\label{fig:memory}
\end{figure}

\paragraph{Language Models for Dialogue}
Many language models have been developed as dialogue agents for chatting with humans. As an early trial in industries, DialoGPT \cite{zhang-etal-2020-dialogpt} and BlenderBot \cite{roller-etal-2021-recipes} employed crawled conversational data to fine-tune pretrained language models (e.g., GPT-2 \cite{radford2019language}) for open-domain dialogue. Built upon an LLM, i.e., GPT-3 \cite{brown2020language}, ChatGPT \cite{openai2022chatgpt} has astounded the community with its powerful chat ability, which is optimized with instruction tuning and alignment tuning. In academia and open-source communities, there have emerged a variety of notable chat language models, such as Vicuna \cite{vicuna2023}, Koala \cite{koala_blogpost_2023}, Baize \cite{xu-etal-2023-baize}, and UltraLLaMA \cite{ding-etal-2023-enhancing}. They are fine-tuned from an LLM named LLaMA \cite{touvron2023llama} using different collected dialogue datasets. Similar efforts are observed in ChatGLM series \cite{du-etal-2022-glm,chatglm-6b-2023,chatglm2-6b-2023} and \textsc{Llama 2}-Chat \cite{touvron2023llama2}. Tuning language models has become a prevalent paradigm for building capable dialogue agents, and this work mainly focuses on open-source LLMs for dialogue.

\paragraph{Consistency in Dialogue}
Dialogue consistency measures whether an agent's generated utterances are consistent with the agent's role and dialogue context, especially from several distinguishable aspects such as topics, styles \cite{wang-etal-2017-steering}, personas \cite{zhang2019consistent,song2020generating,ju-etal-2022-learning}, and characters or roles \cite{urbanek-etal-2019-learning,shuster-etal-2022-state,chen-etal-2023-large}. For checking dialogue consistency, most prior works leveraged natural language inference (NLI) techniques \cite{song2020generating,nie-etal-2021-like} or dataset benchmarking \cite{qin-etal-2021-dont}. 

Existing works have attempted to build persona-consistent dialogues \cite{liu-etal-2020-impress,kim-etal-2020-will,chen2023learning}. For example, \citet{kim-etal-2020-will} adopted the Rational Speech Acts framework to improve persona consistency. Another line of research exhibits that interlocutor modeling is of high necessity for pragmatic communications \cite{bao-etal-2022-learning} and multi-party conversations \cite{gu2023learning}. These studies move a step towards improving dialogue consistency. Nevertheless, the challenge of improving consistency is far from being conquered, even for LLMs \cite{touvron2023llama2}. More recently, MemoChat \cite{lu2023memochat} enhanced LLMs' chat consistency by carefully designing tailored tuning instructions. In comparison, our work is the first to explore how a new way of tuning brings consistency improvement.

\paragraph{Parameter-efficient Tuning}
Conventional fine-tuning is inefficient as the parameter size grows since it requires training all parameters of LLMs. Parameter-efficient tuning \cite{houlsby2019parameter,lester-etal-2021-power} adds a small number of tunable parameter layers, namely \textit{adapters}, for fine-tuning while freezing the original parameters. Prefix Tuning \cite{li-liang-2021-prefix} fine-tunes a sequence of task-specific vectors inserted before the input. LoRA \cite{hu2022lora} adopts trainable low-rank decomposition matrices into LLMs' layers, making it adaptive to new data while retaining the previous knowledge. As LoRA has been widely verified as effective in fine-tuning LLMs and achieving superior performance, this work follows this affordable and reproducible way to develop an efficient tuning framework for multi-round dialogues.

\section{Our Method}
We first provide the necessary background about general dialogue generation and LoRA-based tuning (see \S\ref{sec:preliminaries}). Then, we dive into the details of the proposed \textbf{M}ulti-round \textbf{I}nteractive \textbf{Di}alogue Tuning (\textsc{Midi}-Tuning) framework (see \S\ref{sec:midi_tuning}).

\subsection{Preliminaries}
\label{sec:preliminaries}

\paragraph{Dialogue Generation}
We consider a dialogue dataset as $\mathcal{D}=\{(\mathcal{I}_{i},\mathcal{C}_{i})\}_{i=1}^{N}$ for downstream tasks, where $N$ is the total number of dialogues. $\mathcal{I}_{i}$ denotes task-specific dialogue instruction and necessary additional information, such as domain knowledge facts, specified character descriptions, etc. $\mathcal{C}_{i}=\{<u_{i,t},s_{i,t}>\}_{t=1}^{T}$ denotes utterances between the user $u_{i}$ and agent $s_{i}$ in the $i$-th dialogue, $T$ denotes the total number of dialogue rounds.

Given a task-specific dialogue instruction $\mathcal{I}$ that provides necessary information and a dialogue context $\mathcal{C}=\{<u_1,s_1>,<u_2,s_2>,\cdots,<u_t,>\}$, the objective is to generate a proper agent utterance $s_t$. Essentially, the probability distributions over the agent's utterances are estimated as follows:
\begin{equation}
    p=\prod_{t=1}^{T}p(s_t|s_{<t};u_{\leq t};\mathcal{I})
\end{equation}
More generally, if let $X=[s_{<t};u_{\leq t};\mathcal{I}]$ denote the input context and $Y$ denote the output utterance, the language models-based tuning is to minimize the negative log-likelihood as follows:
\begin{equation}
    \mathcal{L}(\theta)= -\sum_{i=1}^{N}p(Y^{(i)})\log p_{\theta}(\hat{Y}^{(i)}|X^{(i)})
\label{eq:loss}
\end{equation}
where $\theta$ denotes all trainable parameters.

\paragraph{Low-Rank Adaptation}
Low-Rank Adaptation (LoRA) \cite{hu2022lora} hypothesizes that the weight updates in pretrained language models possess a low ``intrinsic rank'' during adaptation. For a pretrained weight matrix $W\in\mathbb{R}^{d\times k}$, it is updated with a low-rank decomposition $W+\Delta W=W+BA$, where $B\in\mathbb{R}^{d\times r}$, $A\in\mathbb{R}^{r\times k}$, and the rank $r\ll\text{min}(d,k)$. During fine-tuning, $W$ is frozen with no gradient updates, while $A$ and $B$ are trainable, making LoRA tuning is much more efficient than full fine-tuning. In practice, LoRA can be specified to adapt the attention weights $W_q$, $W_k$, $W_v$, and $W_o$ corresponding to LLMs' query, key, value, and output projections.

\subsection{\textsc{Midi}-Tuning}
\label{sec:midi_tuning}
We propose the MIDI-Tuning framework, which enables the agent and user to achieve round-level interactions. Figure \ref{fig:framework} shows the overview of our framework. Below, we introduce how it works, from intuitive ideas to technical details.

\paragraph{User-Agent Separate Modeling}
As we emphasized before, it is more feasible to model the roles of the agent and user separately, such that the agent and user models can consistently adhere to their respective roles. As shown in Figure \ref{fig:framework}, we employ an agent adapter like LoRA \cite{hu2022lora} built upon an LLM, e.g., LLaMA \cite{touvron2023llama}, to model the dialogue instruction for the agent (denoted as $s_{\text{inst}}$) and the agent's utterances $s_t$ ($t\geq 1$). Here, $t$ denotes the dialogue round. We employ another LoRA adapter built upon the same LLM to model the user's utterances $u_t$. The backbone LLM is shared since it is frozen during tuning, while the weight parameters of the two LoRA adapters are trainable to distinguish role disparities between the agent and user. 

Mathematically, we decompose the probability distribution over all the utterances in dialogue into two distributions for the \textit{user model} and \textit{agent model}, respectively, shown as follows:
\begin{align}
    p(u)&=\prod_{t=1}^{T}p(u_t|u_{<t};s_{<t}) \\
    p(s)&=\prod_{t=1}^{T}p(s_t|s_{<t};u_{\leq t};\mathcal{I})
\end{align}
where $p(u)$ and $p(s)$ are language models whose task is to predict the next token given the preceding context. Finally, the objective of our tuning is to optimize the joint losses of the agent model (denoted as $\mathcal{L}_{s_t}$) and user model (denoted as $\mathcal{L}_{u_t}$):
\begin{equation}
    \mathcal{L}=\mathcal{L}_{s_t} + \beta\mathcal{L}_{u_t}
\end{equation}
where $\beta$ is a hyperparameter controlling the weight.

\paragraph{Round-level Memory Caching}
One of the key challenges of the introduced separate modeling is that tracking the complete dialogue context is non-trivial. To address it efficiently, we propose a round-level memory caching mechanism in this section.
Since Transformer \cite{vaswani2017attention} lays the foundation architecture of existing LLMs, it is capable of applying memory recurrence \cite{dai-etal-2019-transformer,wang-etal-2020-dual,wu-etal-2021-alternating} to cache Transformer's self-attention ($Q$, $K$, $V$ for queries, keys and values) computations to maintain context information during user-agent interactions. 

Let us define $h_t$ as the hidden states for an utterance at $t$-th round, $M_{t}$ as the memory at $t$-th round containing all key-value pairs from the past. As shown in Figure \ref{fig:framework} and Figure \ref{fig:memory}, we reuse history keys (i.e., $K_{\leq t-1}$) and values (i.e., $V_{\leq t-1}$) as the cached memory $M_{t-1}$, to perform self-attention computation to obtain $h_t$, and then store $h_t$ back to the memory as $M_t$. The keys and values computed from previous rounds are fixed and cached to be reused as ongoing context when the agent/user model processes the present-round utterance (see Figure \ref{fig:memory}), allowing the model to exploit information in history. The entire process is formulated as follows:
\begin{align}
    M_{t}&=[(K_{\leq t}^{(1)},V_{\leq t}^{(1)}),\cdots,(K_{\leq t}^{(l)},V_{\leq t}^{(l)})] \\
    K_{\leq t}^{(i)} &=[K_{\leq t-1}^{(i)};K_{t}^{(i)}] \\
    V_{\leq t}^{(i)} &=[V_{\leq t-1}^{(i)};V_{t}^{(i)}] \\
    h_{t}^{(i)}&=\text{Attention}(Q_{t}^{(i)}, K_{\leq t-1}^{(i)}, V_{\leq t-1}^{(i)})
\end{align}
where $[\cdot;\cdot]$ denotes concatenation, $h_{t}^{(i)}$ is the hidden states at the $i$-th  layer. The last layer's hidden states $h_t$ is used to calculate loss during tuning and to obtain generation probability during inference.

Since there are two LoRA adapters, two individual $W_q$ weight matrices for query projections will be trainable. Recall that obtaining a good agent model is the ultimate goal, we adopt a \textit{context value protection} strategy to train the agent model's value projection, i.e., the weight matrix $W_v$ of LoRA, without training the user model's value projection. This operation enables the agent model to exploit context value in a consistent space.

\paragraph{Tuning and Inference}
Although the idea presented before is appealing, some technical challenges still need to be solved in practice. \textbf{First}, the rounds of different dialogues and sequence lengths of different utterances within one dialogue might be unequal, \textit{how can we achieve batched tuning on downstream data}? We pad batched utterances to the maximum utterance length within a batch, and similarly, pad instructions to the maximum batched instruction length. We sort batched dialogues by their rounds in descending order, similar to inverted triangular causal masks, making it easier to compute losses for valid utterances. We set a maximum number of rounds according to downstream tasks, truncating early-round utterances if longer.

\textbf{Second}, the paddings among different rounds result in the positions of utterance tokens not continuous since most LLMs adopt the Rotary Position Embedding \cite{su2021roformer}. When reusing the cached memory, \textit{how can we keep the positional information consecutive}? To this end, we set valid positional ids at each round by counting valid tokens and masking out positions that should not be seen. Then, we explicitly pass the necessary positional ids as part of the model input during both training and inference.

Our inference process differs from that of traditional methods. We feed the past ground-truth utterances $<u_{\leq t},s_{<t}>$ round by round to obtain the cached memory, which is finally used to generate the agent's corresponding utterance $s_{t}$ at $t$-th round. In realistic interactions, we use the memory yielded from previously generated utterances since we do not have ground-truth dialogue history.

\section{Experiments}

\subsection{Experimental Setting}

\paragraph{Tasks}
We consider validating our framework on two challenging dialogue tasks: \textit{character-based dialogue} \cite{urbanek-etal-2019-learning,han-etal-2022-meet,chen-etal-2023-large} and \textit{target-oriented proactive dialogue} \cite{wang-etal-2023-target,wang-etal-2023-dialogue,deng-etal-2023-prompting}. For character-based dialogue, the challenge for an agent lies in maintaining character identity consistent with the assigned role throughout the conversation, where the agent may incorrectly take on the roles or activities of its faced users \cite{shuster-etal-2022-state} instead of its assigned role. For target-oriented proactive dialogue, an agent should proactively direct the conversation towards its assigned target (a specific goal) step by step. This long-term goal-directed behavior makes it non-trivial to maintain the consistency that adheres to its goal with the dialogue rounds moving forward.

\begin{table}[t!]
\centering
\resizebox{1.0\linewidth}{!}{
\begin{tabular}{lcccc}
\toprule
 & Train  & Valid & Test-Seen  & Test-Unseen   \\
\midrule
\# characters & 934 & 410  & 593  & 292  \\
\# dialogues & 8,307 & 500  & 1,000  & 721 \\
\# utterances & 110,265 &  6,654  & 13,392  &  9,818 \\
\# utterances / dialogue  &  13.3  & 13.3  & 13.4  &  13.6 \\
\bottomrule
\end{tabular}}
\caption{Statistics of the \textsc{Light} dataset.}
\label{tab:stat_light}
\end{table}

\begin{table}[t!]
\centering
\resizebox{1.0\linewidth}{!}{
\begin{tabular}{lc}
\toprule
\# dialogues (Train / Valid / Test) & 12,601 / 1,802 / 3,606 \\
\# utterances (Train / Valid / Test) &  141,928 / 20,310 / 40,496  \\
Total \# targets  &   501  \\
Avg. \# utterances / dialogue  &  12.3 \\
\bottomrule
\end{tabular}}
\caption{Statistics of the \textsc{TopDial} dataset.}
\label{tab:stat_topdial}
\end{table}

\paragraph{Datasets}
Our experiments are conducted on the \textbf{\textsc{Light}} \cite{urbanek-etal-2019-learning} and \textbf{\textsc{TopDial}} \cite{wang-etal-2023-target} datasets. \textsc{Light} is a character-based dialogue dataset collected from crowdworker interactions with a set of game location settings (e.g., countryside, forest, castle). It contains various game characters, from animals to humans (e.g., dragon, wizard, servant). Each dialogue has a background description of the setting, while each character has a persona with several sentences describing its traits (see detailed examples in Appendix \ref{appendix:light}). Table \ref{tab:stat_light} shows statistics of \textsc{Light}, where the \textit{seen} test set consists of dialogues with their locations and characters that can appear in the training set. In contrast, the \textit{unseen} test set comprises dialogues collected on the unseen set of locations and characters, providing a more challenging test. 

\textsc{TopDial} is a target-oriented dialogue dataset for proactive agents with personalized users. The agent is assigned a target consisting of a <dialogue act, topic> pair, where these target dialogue acts mainly lie in recommendations on the domains of movies, music, and food. The agent must proactively lead the discussed topic towards the target topic based on domain knowledge, and meanwhile, adapt its faced user's personalized aspects (e.g., profiles and personalities) to maintain engagement instead of obtrusively driving to the target. The agent's ultimate task is to achieve the target act on the target topic (see detailed examples in Appendix \ref{appendix:topdial}). Table \ref{tab:stat_topdial} shows statistics of \textsc{TopDial}. Appendix \ref{appendix:dataset} describes more details for preprocessing the data into the general format with instructions.

\subsection{Evaluation}

\paragraph{Consistency Evaluation}
Consistency evaluation in dialogue has been a long-standing yet challenging problem \cite{nie-etal-2021-like,shuster-etal-2022-state,han-etal-2022-meet}. Inspired by these prior studies, we utilize a binary classifier trained on the downstream datasets to measure the consistency probability (\textbf{Consist. Prob.}) of the agent's generated responses. We concatenate a given context input and a response as the complete input to yield the classification label $y\in\{1(\text{consistent}),0 (\text{inconsistent})\}$.
For each ground-truth (consistent) response in the \textsc{Light} dataset, we construct the inconsistent set by sampling from (1) the user's utterances under the current dialogue setting due to the user's character being obviously different from the agent's, and (2) the agent's utterances with the same character but under different dialogue settings. Then, we fine-tune a pretrained BERT \cite{devlin-etal-2019-bert} model followed by a linear layer for binary classification, producing an automatic consistency estimator. We also employ this method to train a consistency estimator on the \textsc{TopDial} dataset accordingly. Appendix \ref{appendix:consist_estimator} provides complete details.

\begin{figure}[t!]
\centering
\subfigure[ROC Curve]{
    \begin{minipage}[b]{0.465\linewidth}
	    \includegraphics[width=1\linewidth]{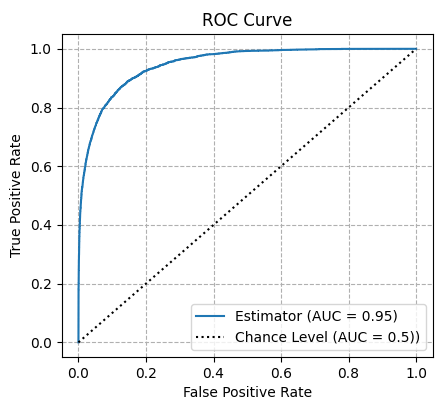}
    \end{minipage}
    \label{fig:consist_roc}
}
\subfigure[Calibration Curve]{
    \begin{minipage}[b]{0.465\linewidth}
	    \includegraphics[width=1\linewidth]{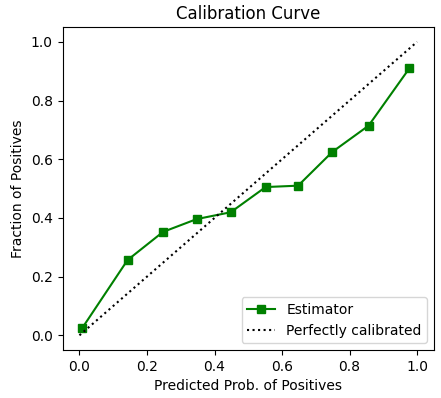}
    \end{minipage}
    \label{fig:consist_calib}
}
\caption{Performance of the created consistency estimator on the \textsc{Light} validation set.}
\label{fig:consist_curve}
\end{figure}

Figure \ref{fig:consist_curve} shows the performance of our trained consistency estimator on the \textsc{Light} validation set. The convex ROC curve (see Figure \ref{fig:consist_roc}) with an AUC (Area Under the Curve) value of 0.95 shows that our estimator is highly discriminative in recognizing whether an agent's response is consistent with the dialogue context. Meanwhile, the calibration curve (see Figure \ref{fig:consist_calib}) indicates that our estimator has high confidence in its predicted probabilities for positive (i.e., consistent) responses. Therefore, it is reliable to use our trained estimator to automatically measure the consistency of an agent's generated response during evaluation.

In addition, we adopt the state-of-the-art LLM, GPT-4 \cite{openai2023gpt4}, to automatically evaluate the score of dialogue consistency (\textbf{GPT-4 Score}), similar to existing works \cite{zheng2023judging,lu2023memochat}. We take the necessary checking information (e.g., specified character descriptions), dialogue context, and the agent's generated response as a whole, then ask GPT-4 to rate the consistency with an integer scale of 1 $\sim$ 10. We provide the details of the prompt setting in Appendix \ref{appendix:consist_gpt4}.

\paragraph{Dialogue Evaluation Metrics}
In addition to consistency, we also adopt commonly used automatic evaluation metrics for dialogue generation. Our evaluation metrics include word-level F1 (\textbf{Word F1}), \textbf{BLEU}-$n$ \cite{papineni-etal-2002-bleu}, and distinct (\textbf{DIST}) \cite{li-etal-2016-diversity} for the \textsc{Light} dataset. For the \textsc{TopDial} dataset, we adopt the \textbf{Word F1}, \textbf{BLEU}-$n$, and target success rate (\textbf{Succ.}) \cite{wang-etal-2023-target}, following prior studies \cite{wang-etal-2023-target,dao-etal-2023-reinforced} for target-oriented proactive dialogue. Appendix \ref{appendix:eval_metrics} provides the details of the above metrics.

\subsection{Implementation}

\paragraph{Baseline Models}
We adopt popular open-source LLMs as baseline models for experiments, including \textbf{LLaMA} \cite{touvron2023llama}, \textbf{Mistral-7B} \cite{jiang2023mistral}, \textbf{Vicuna} \cite{vicuna2023}, and \textbf{\textsc{Llama 2}-Chat} \cite{touvron2023llama2}. As our primary focus is the way of tuning, we mainly consider using 7B-size models since they are widely compute-affordable. Our framwork can be easily adapted to much larger models, e.g., with a size of 13B or 70B.

\begin{table*}[t!]
\centering
\resizebox{0.98\textwidth}{!}{
\begin{tabular}{clccccc}
\toprule
  & \textbf{Model} & \textbf{Consist. Prob.} & \textbf{GPT-4 Score} & \textbf{Word F1 (\%)}  & \textbf{BLEU-1 / 2}  & \textbf{DIST-1 / 2} \\
\midrule 
\multirow{3}{*}{No Tuning} & GPT-3.5-Turbo  & 0.653 &  7.23 &  18.05  & 0.137 / 0.049  & 0.026 / 0.206  \\
\cdashline{2-7}
 & LLaMA-7B & 0.378 & 4.22 &  12.20 & 0.074 / 0.025 & 0.016 / 0.112  \\
  & Mistral-7B &  0.528  &  6.80   & 13.51    &  0.099 / 0.037 & 0.021 / 0.131  \\
 & \textsc{Llama 2}-Chat-7B & 0.535 & 6.73 & 14.98 &  0.095 / 0.030 & 0.023 / 0.177 \\
 & Vicuna-7B &  0.620 & 6.85 & 20.54 & 0.145 / 0.051 & \textbf{0.040} / \textbf{0.257} \\
\midrule 
\multirow{3}{*}{Fine-tuning (FT)} & LLaMA-7B & 0.449 & 4.86 & 18.62 & 0.122 / 0.042 & 0.037 / 0.223 \\
 & Mistral-7B & 0.611   &  7.05 &  20.19   & 0.140 / 0.053 & 0.036 / 0.204    \\
 & \textsc{Llama 2}-Chat-7B & 0.584 & 6.88 & 20.09 & 0.134 / 0.051 & 0.035 / 0.202 \\
 & Vicuna-7B & 0.650 & 7.32 & 20.51 & \textbf{0.145} / 0.056  & 0.036 / 0.208 \\
\midrule
\multirow{3}{*}{\textbf{\textsc{Midi}-tuning (Ours)}} & LLaMA-7B & 0.563 ($\uparrow$ \textbf{25.4 \%}) & 5.52 ($\uparrow$ \textbf{13.6\%})
 & 19.68 & 0.125 / 0.049  & 0.037 / 0.198 \\
  & Mistral-7B &  0.626 ($\uparrow$ 2.5 \%)  &  7.40 ($\uparrow$ 5.0 \%)     &  20.22  &  0.141 / 0.055 & 0.036 / 0.206   \\
 & \textsc{Llama 2}-Chat-7B & 0.635 ($\uparrow$ 8.7\%)  & 7.46 ($\uparrow$ 8.4\%) & 20.27 & 0.132 / 0.051 & 0.038 / 0.209 \\
 & Vicuna-7B & \textbf{0.657} ($\uparrow$ 1.1\%)  & \textbf{7.65} ($\uparrow$ 4.5\%) & \textbf{20.56}  & 0.140 / \textbf{0.057} & 0.038 / 0.213 \\
\bottomrule
\end{tabular}}
\caption{Automatic evaluation results of dialogue generation on the \textsc{Light} test-\textit{seen} set ($\uparrow$ denotes ours v.s. FT).}
\label{tab:result_light_seen}
\end{table*}

\begin{table*}[t!]
\centering
\resizebox{0.98\textwidth}{!}{
\begin{tabular}{clccccc}
\toprule
  & \textbf{Model} & \textbf{Consist. Prob.} & \textbf{GPT-4 Score} & \textbf{Word F1 (\%)}  & \textbf{BLEU-1 / 2}  & \textbf{DIST-1 / 2} \\
\midrule 
\multirow{3}{*}{No Tuning} & GPT-3.5-Turbo  & 0.636  &  7.02  &  17.83   & 0.128 / 0.046  & 0.028 / 0.216  \\
\cdashline{2-7}
 & LLaMA-7B &  0.390 & 4.32 & 11.30 & 0.067 / 0.023 & 0.017 / 0.114 \\
 & Mistral-7B &  0.548   &  6.15  &  12.78   & 0.096 / 0.036 & 0.020 / 0.120  \\
 & \textsc{Llama 2}-Chat-7B & 0.496 & 5.88 & 13.91 & 0.087 / 0.027 & 0.026 / 0.186 \\
 & Vicuna-7B & 0.614 & 6.85 & 19.18 & 0.142 / 0.053 &  0.041 / \textbf{0.267} \\
\midrule 
\multirow{3}{*}{Fine-tuning (FT)} & LLaMA-7B & 0.445 & 5.34 & 18.15 & 0.123 / 0.042 & 0.041 / 0.221  \\
 & Mistral-7B & 0.605  &  6.96  &  20.18  &  0.140 / 0.052  & 0.039 / 0.217  \\
 & \textsc{Llama 2}-Chat-7B & 0.570 & 6.69  & 20.15 & 0.142 / \textbf{0.055} & 0.039 / 0.214 \\
 & Vicuna-7B & 0.646 & 7.11 & 20.26 & \textbf{0.144} / \textbf{0.055} & 0.039 / 0.219  \\
\midrule
\multirow{3}{*}{\textbf{\textsc{Midi}-tuning (Ours)}} & LLaMA-7B & 0.559 ($\uparrow$ \textbf{27.6\%}) & 6.02 ($\uparrow$ \textbf{12.7\%}) & 19.70 & 0.128 / 0.050  &  \textbf{0.042} / 0.212 \\
 & Mistral-7B &  0.621 ($\uparrow$ \textbf{2.6\%}) &   7.16 ($\uparrow$ \textbf{2.9\%}) & 20.30  & 0.139 / 0.052    &  0.040 / 0.218  \\
 & \textsc{Llama 2}-Chat-7B & 0.620 ($\uparrow$ 8.8\%) & 7.19 ($\uparrow$ 7.5\%) &  20.28  & 0.137 / 0.053 & 0.040 / 0.219 \\
 & Vicuna-7B & \textbf{0.664} ($\uparrow$ 2.8\%) & \textbf{7.40} ($\uparrow$ 4.1\%) & \textbf{20.72} & 0.140 / \textbf{0.055} & 0.041 / 0.222 \\
\bottomrule
\end{tabular}}
\caption{Automatic evaluation results of dialogue generation on the \textsc{Light} test-\textit{unseen} set ($\uparrow$ denotes ours v.s. FT).}
\label{tab:result_light_unseen}
\end{table*}

\paragraph{Baseline Settings}
We consider the following two settings for all baseline models: (\romannumeral1) \textbf{\textit{No Tuning}}, which indicates that each model directly takes the concatenated text of the task instruction and a dialogue context as input prompt, then generates an utterance as the agent's response. Since this setting performs without any tuning, it can be used to measure the fundamental chat ability of an LLM and the difficulty level of a downstream dialogue task. We also include the GPT-3.5-Turbo version of ChatGPT as an additional baseline for this setting.  (\romannumeral2) \textbf{\textit{Fine-tuning}}, which tunes an LLM based on LoRA \cite{hu2022lora} using the downstream training set, following the conventional one-dialogue-one-sample tuning manner since the adopted baseline models are all causal LLMs.

Note that we ensure the input task instructions of the \textit{No Tuning}, \textit{fine-tuning}, and \textit{ours} are identical for a test sample, following the format as described in Appendix \ref{appendix:dataset}. This will mitigate the influence of generation caused by different instructions. We have two additional special tokens, e.g., \texttt{[USER]} and \texttt{[ASSISTANT]}, inserted ahead of each utterance from the user and agent, respectively.

\begin{figure}[t!]
\centering
\includegraphics[width=0.94\linewidth]{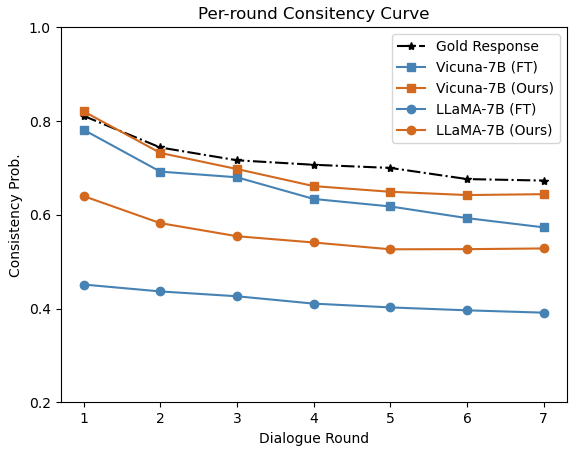}
\caption{Per-round consistency comparison between the fine-tuning (FT) and \textsc{Midi}-Tuning (Ours) on the \textsc{Light} test-\textit{unseen} set.}
\label{fig:consist_comp}
\end{figure}

\begin{table*}[t!]
\centering
\resizebox{0.93\textwidth}{!}{
\begin{tabular}{clccccc}
\toprule
  & \textbf{Model} & \textbf{Consist. Prob.} & \textbf{GPT-4 Score}  & \textbf{Word F1 (\%)}  &  \textbf{BLEU-1 / 2}  & \textbf{Succ. (\%)} \\
\midrule 
\multirow{3}{*}{No Tuning} & GPT-3.5-Turbo  & 0.806   & 8.33    &  42.06   &  0.348 / 0.237 &  65.22 \\
\cdashline{2-7}
 & LLaMA-7B & 0.550 &  6.35  & 34.28 & 0.280 / 0.155 & 35.89 \\
  & Mistral-7B &  0.655  &  7.64    &   31.04  & 0.215 / 0.116 & 39.06   \\
  & \textsc{Llama 2}-Chat-7B & 0.686 & 7.62  & 35.70 & 0.292 / 0.160 & 39.54  \\
 & Vicuna-7B & 0.632  &  7.58  &  36.84 & 0.305 / 0.171 & 41.55  \\
\midrule 
\multirow{3}{*}{Fine-tuning (FT)} & LLaMA-7B &  0.771 & 8.28 &  40.64 & 0.311 / 0.203 & 65.56  \\
 & Mistral-7B & 0.794   &  8.50  &  45.08  & \textbf{0.396} / \textbf{0.271} &  68.64 \\
 & \textsc{Llama 2}-Chat-7B &  0.793  & 8.12 & 42.89 & 0.341 / 0.223 & 73.97 \\
 & Vicuna-7B & 0.821 & 8.55  &  44.59 & \textbf{0.396} / 0.264 & 75.40 \\
\midrule
\multirow{3}{*}{\textbf{\textsc{Midi}-tuning (Ours)}} & LLaMA-7B & 0.796 ($\uparrow$ \textbf{3.2\%}) & 8.40 ($\uparrow$ \textbf{1.4\%}) &  42.50 & 0.336 / 0.210 & 66.89 \\
 & Mistral-7B &  0.813 ($\uparrow$ 2.4\%) & 8.59 ($\uparrow$ 1.1\%)  & 44.36    & 0.392 / 0.270 & 70.15   \\
 & \textsc{Llama 2}-Chat-7B & 0.815 ($\uparrow$ 2.8\%) & 8.20 ($\uparrow$ 1.0\%) & 43.52 & 0.355 / 0.225 & 72.20 \\
 & Vicuna-7B &  \textbf{0.836} ($\uparrow$ 1.8\%)  & \textbf{8.65} ($\uparrow$ 1.2\%)  &  \textbf{45.40} & \textbf{0.396} / \textbf{0.271} & \textbf{76.07} \\
\bottomrule
\end{tabular}}
\caption{Automatic evaluation results of dialogue generation on the \textsc{TopDial} test set ($\uparrow$ denotes ours v.s. FT).}
\label{tab:result_topdial}
\end{table*}

\paragraph{Implementation Details}
\label{sec:implement}
We implement the baseline settings and our \textsc{Midi}-Tuning using the Huggingface PEFT \cite{peft} library, and we incorporate DeepSpeed \cite{rasley2020deepspeed} to improve the training efficiency. For all experiments that involve tuning, the LoRA's target modules are $W_q$ and $W_v$, the rank $r$ is set to 8, and the scaling parameter $\alpha$ is set to 16. We adopt 4-bit quantization \cite{dettmers2023qlora} for efficient finetuning of LLMs. The optimizer we used is AdamW \cite{loshchilov2018decoupled}, with a warmup ratio of 0.03. The learning rate is set to 2$e$-5 with a cosine scheduler. Due to the memory constraint, the maximum number of dialogue rounds is set to 10.
The maximum text window for all models is set to 2k, sufficient to cover the context length for the two datasets. The hyperparameter $\beta$ is set in the range ($0,1$].
The other hyperparameters are set as the default, following Vicuna \cite{vicuna2023}. Appendix \ref{appendix:implement} provides more details on tuning and inference.

\section{Results and Discussions}

\subsection{Automatic Evaluation Results}
Table \ref{tab:result_light_seen} and Table \ref{tab:result_light_unseen} report the automatic evaluation results on the \textsc{Light} test-\textit{seen} and test-\textit{unseen} datasets, respectively. Though GPT-3.5-Turbo performs very well, we find that the majority of baseline models with no tuning perform inferior, indicating that merely relying on prompting may not be effective enough for the \textsc{Light}-like dialogue tasks. With downstream training data, vanilla fine-tuning enables these baseline models to deeply understand a specific dialogue task and achieve much better generation performances.
Nonetheless, our \textsc{Midi}-Tuning outperforms vanilla fine-tuning remarkably in terms of the consistency probability and GPT-4 score, and meanwhile, achieves higher or on par with scores in other dialogue generation metrics (e.g., word F1, BLEU). We observe a similar trend between the consistency probability predicted by our created estimator and the GPT-4 score rated by GPT-4, widely demonstrating the effectiveness of our framework in consistency improvement. Similarly, as shown in Table \ref{tab:result_topdial}, our \textsc{Midi}-Tuning performs better than vanilla fine-tuning on the \textsc{TopDial} dataset. 

Overall, our \textsc{Midi}-Tuning is superior in generating more consistent responses without compromising much performance in other aspects.

\subsection{Per-round Consistency Analysis}
To look at how our \textsc{Midi}-Tuning performs as the dialogue rounds moving forward, we visualized per-round consistency comparison between the fine-tuning (FT) and ours on the \textsc{Light} test-\textit{unseen} set. Figure \ref{fig:consist_comp} shows the comparison results, where the curve for gold response is obtained by feeding ground-truth response at each round into the created consistency estimator, serving as an approximal upper bound for per-round consistency.

As shown in Figure \ref{fig:consist_comp}, LLaMA-7B with our \textsc{Midi}-Tuning achieves a large margin of consistency improvement compared to that with FT. We highlight the importance of our framework for promoting consistent dialogue generation since LLaMA is a foundation LLM with neither instruction tuning nor human alignment, which can be a fair pedestal for comparing different tuning methods. We observe that Vicuna-7B with FT performs closely compared to Vicuna-7B with ours, which might be attributed to Vicuna's intrinsic powerful chat ability since it is an instruction-tuned LLM based on a variety of dialogue data. More importantly, the estimated consistency of both LLaMA-7B and Vicuna-7B with our \textsc{Midi}-Tuning drops slowly and even maintains stable (e.g., 5 $\sim$ 7 rounds), while the estimated consistency is continuously declining with the dialogue rounds going on for the two models with vanilla fine-tuning. It suggests that the proposed method is able to maintain multi-round dialogue consistency.

\subsection{Human Evaluation}

To further assess the performance of the proposed framework in realistic multi-round dialogue scenarios, we conducted an interactive evaluation following existing studies \cite{li2023camel,cheng2024cooper}. We used ChatGPT to simulate the roles of different users based on the \textsc{TopDial} test set and converse with the evaluated models round by round, where we adopted LLaMA and Vicuna, tuned by vanilla fine-tuning (FT) and our \textsc{Midi}-Tuning. Given a pair of dialogues produced by two variants of tuning for the same backbone model, we recruited three well-educated graduate students as the annotators to evaluate which one is better (or tied) from the dimensions of consistency (\textbf{Consist.}), proactivity (\textbf{Proact.}), coherence (\textbf{Coh.}), and target success rate (\textbf{Succ.}). We provide the metrics explanations and detailed setting in Appendix \ref{appendix:human_eval}.

\begin{figure}[t!]
\centering
\includegraphics[width=1.0\linewidth]{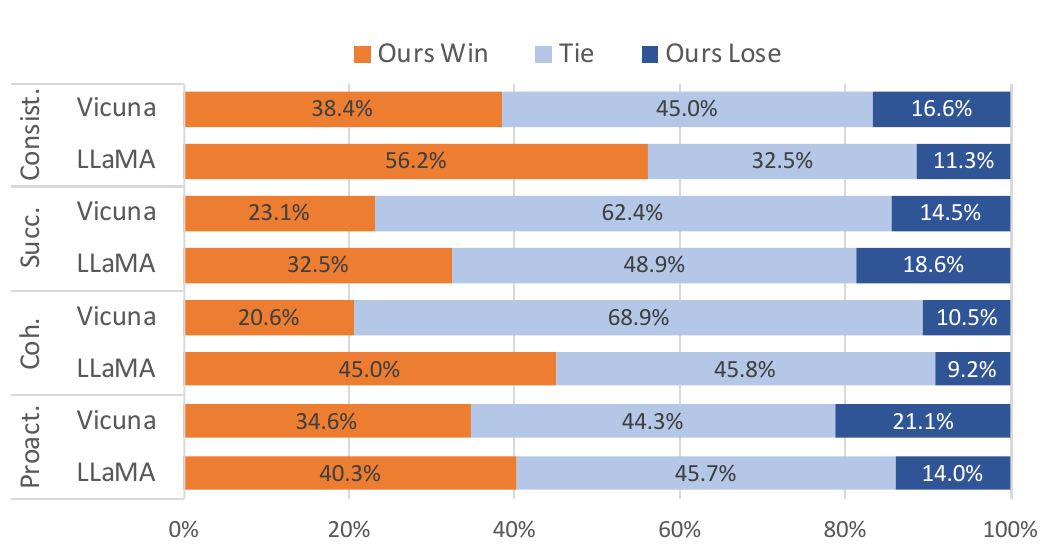}
\caption{Human evaluation results of the fine-tuning (FT) and \textsc{Midi}-Tuning (Ours).}
\label{fig:human_eval}
\end{figure}

Figure \ref{fig:human_eval} shows the comparison results between FT and ours. We obtain an average Fleiss's kappa of $\kappa$ = 0.486, indicating a moderate (0.41 < $\kappa$ < 0.60) agreement among annotators. The results shown in Figure \ref{fig:human_eval} suggest that our \textsc{Midi}-Tuning significantly outperforms FT in maintaining consistency (56.2\% win rate for LLaMA) in the interactive setting. It is also superior to or on par with FT in other dimensions. To give a better sense of generation quality, we provide a case study in Appendix \ref{appendix:case_study}.
In summary, our \textsc{Midi}-Tuning is more effective in generating consistent, coherent, and appropriate utterances.

\section{Conclusion}
This work explores how the way of tuning can improve the consistency of dialogue generation over multiple rounds. We highlight the importance of separately modeling agents and users due to their role disparities. We propose a general, efficient tuning framework called \textsc{Midi}-Tuning, which represents the agent and user using two adapters and tunes them via round-level memory caching. Empirical experiments show that our framework outperforms traditional dialogue tuning significantly.

\section*{Limitations}
We recognize the limitations of this work in the following aspects. First, our \textsc{Midi}-Tuning framework requires padding among dialogue rounds to achieve batched tuning. It might result in redundant GPU memory consumption (see Appendix Table \ref{tab:memory}) as the rounds become longer and longer. 
Second, the current framework needs to be more compute-efficient since it cannot compute losses in parallel for different rounds of utterances. 
We will consider improving the compute efficiency of our framework by employing advanced acceleration techniques, such as \textsc{FlashAttention} \cite{dao2022flashattention}.
Third, the \textsc{Midi}-Tuning relies on the architecture of causal language models for encoding, decoding, and round-level memory caching. It cannot directly tune encoder-decoder language models for dialogue generation. We will leave addressing these challenges as our future work.

\section*{Ethics Statement}
This work mainly focuses on developing a general, efficient framework to tune LLMs for multi-round dialogue generation. The adopted LLMs for tuning are all open-sourced. We strictly follow the protocols for the academic use of these LLMs. Our experimental datasets are publicly available and do not involve sensitive or private information. It is also known that response generation from these LLMs may have concerns about toxicity and bias. Thus, we emphasize that ensuring safe deployment and interaction is a necessity. In addition, we partially use AI assistants, such as Copilot and ChatGPT, to assist us with coding and writing.

\section*{Acknowledgments}
This work was supported by the National Natural Science Foundation of China (62076212), the Research Grants Council of Hong Kong (15207122, 15207920, 15207821), and PolyU internal grants (ZVQ0, ZVVX). The authors would like to thank the anonymous reviewers for their valuable feedback and constructive suggestions.

\bibliography{custom}
\bibliographystyle{acl_natbib}

\appendix

\section{Dataset Preprocessing}
\label{appendix:dataset}

\paragraph{\textsc{Light} Dataset}
\label{appendix:light}
Figure \ref{fig:example_light} shows an example dialogue from the \textsc{Light} \cite{urbanek-etal-2019-learning} dataset. For each dialogue in the dataset, we transform the given character-related information and the setting description into natural languages, following the instruction template presented in Figure \ref{fig:instruct_light}. As such, we obtain dialogue data with the instruction-based format (a dialogue instruction for the agent and multi-round user-agent utterances), as we introduced in the preliminaries.

\begin{figure*}[th!]
\begin{AIBox}{}
\parbox[t]{0.96\textwidth}{{} 
\small\begin{alltt}
\textbf{Character 1}: \textbf{Servant (User) }   \\
\textbf{Persona Description}: I come from the lower class. I do what I am told without question. I can not read. I have not seen my family in a long time.
\tcbline
\textbf{Character 2}: \textbf{King (Agent)}   \\
\textbf{Persona Description}: I am a king of the whole empire. I give rules and pursuit them. I am brave and fearless. I am carrying a crown and a scepter.
\tcbline
\textbf{Setting}: The main foyer is massive. A grand staircase sits to the back of the foyer leading to the upstairs. At the front of the foyer stand two servants ready to help anyone who comes to visit. To the left of the room there is a doorway leading into a corridor. To the right there is a door leading to another corridor for the King’s servants. At the foot of the stairs there is a bearskin rug that is staring at you almost as if still hungry. The walls are lined with portraits of the king and his family.
\tcbline
\textbf{Servant (User)}: My humble king. What am I to do to serve you? \\
\textbf{King (Agent)}: Ahhh. My loyal servant. Polish my scepter. \\
\textbf{Servant (User)}: Yes my lord. I will polish it immediately. Am I to return it to you personally? \\
\textbf{King (Agent)}: Yes. Yes. Of course. Also check the jewels in my crown. They seem loose. \\
\textbf{Servant (User)}: But sire I am not qualified to do that. Would you prefer I take it to someone? \\
\textbf{King (Agent)}: Oh fine then. \\
\textbf{Servant (User)}: I am sorry sir the rug startled me. \\
\textbf{King (Agent)}: Haha! That’s bear I slain on my latest hunting trip. He’s a mighty beast! \\
\textbf{Servant (User)}: And if I may ask where did you go hunting sire? \\
\textbf{King (Agent)}: The great woods of course. This bear was stealing children in the kingdom. Surely you heard about it. \\
\textbf{Servant (User)}: Sire. I have not been outside of these walls in quiet some time. I have not seen my family in ages. \\
\textbf{King (Agent)}: Such is the life of a servant I suppose. How’s that scepter looking? \\
\textbf{Servant (User)}: It is almost ready sire. and the crown who would you like me to take it to? \\
\textbf{King (Agent)}: Here just give it back. I’ll have the queen find someone.
\end{alltt}}
\end{AIBox}
\caption{An example dialogue from the \textsc{Light} dataset.}
\label{fig:example_light}
\end{figure*}

\begin{figure*}[th!]
\begin{AIBox}{}
\parbox[t]{0.96\textwidth}{
\small\begin{alltt}
Suppose you are a <AGENT\_CHRACTER>. You have the following character persona: \\
\#\# <AGENT\_PERSONA\_DESCRIPTION> \\
\\
You are chatting with a <USER\_CHRACTER>, whose persona is below: \\
\#\# <USER\_PERSONA\_DESCRIPTION> \\
\\
You are chatting with each other under the following setting: \\
\#\# <SETTING> \\
\\
Please write a response that appropriately continues the following conversation.
\end{alltt}}
\end{AIBox}
\caption{Instruction template for the dialogue agent on the \textsc{Light} dataset.}
\label{fig:instruct_light}
\end{figure*}

\paragraph{\textsc{TopDial} Dataset}
\label{appendix:topdial}
Figure \ref{fig:example_topdial} shows an example dialogue from the \textsc{TopDial} \cite{wang-etal-2023-target} dataset. For each dialogue in the dataset, we transform the assigned target, domain knowledge facts, and user information into natural languages, following the instruction template \cite{wang-etal-2023-target} presented in Figure \ref{fig:instruct_topdial}. Similarly, we obtain dialogue data with the instruction-based format, as we introduced in the preliminaries.

\begin{figure*}[th!]
\begin{AIBox}{}
\parbox[t]{0.96\textwidth}{{} 
\small\begin{alltt}
\textbf{Agent's Target}:  <Movie recommendation, ``To Me the Way''>
\tcbline
\textbf{Domain Knowledge}:  <``To Me the Way'', Type, ``Motion Short film>, <``To Me the Way'', Stars, ``Gu Tianle, Jay Chou, Edison Chan, Wu Jianhao''>, ... \\
\textbf{User Profile}:  \{Age Range: ``26-35'', Name: ``Xinqi Ren'', Gender: ``Male'', Residence: ``Dongguan'', Occupation": ``Student'', Accepted movies: ``Days of Tomorrow'', Accepted music: ``The Promised Love; Magician'', Rejected music: ``Rooftop'',  Accepted celebrities: ``Andy Lau; Jay Chou'', Rejected movies: ``''\} \\
\textbf{User Personality}:  \{Openness: ``intellectual, imaginative, and curious'', Conscientiousness: ``inefficient, careless, and sloppy'', Extraversion: ``shy, reserved, and quiet'', Agreeableness: ``trustworthy, straightforward, and generous'', Neuroticism: ``sensitive, nervous, and insecure''\}
\tcbline
\textbf{[Agent]}: Hello Xinqi Ren, it's great to chat with another movie enthusiast! What are some of your favorite movies and music artists? \\
\textbf{[User]}: Hi Yuhang Wang, I enjoy Days of Tomorrow and music from The Promised Love and Magician. My favorite celebrities are Andy Lau and Jay Chou. I'm not a fan of Rooftop though.  \\ 
\textbf{[Agent]}: I see that you're a big fan of Jay Chou. Did you know that he was named one of the 25 Most Influential People in Asia by CNN and was also included in Fast Company's Top 100 Creative People in the World? He's achieved a lot!  \\ 
\textbf{[User]}: Yes, I'm aware of Jay Chou's achievements. He's definitely a talented musician and I enjoy his music. However, I tend to be more reserved and shy, so I don't really follow celebrity news or gossip much. I just appreciate the music and movies. \\ 
\textbf{[Agent]}: That's totally understandable. Speaking of movies, have you watched To Me the Way? It's a motion short film that has a good reputation and features Jay Chou as one of the stars. It's definitely worth checking out!  \\ 
\textbf{[User]}: I haven't seen To Me the Way, but I'll definitely check it out. \\ 
\textbf{[Agent]}: To Me the Way actually has a rating of 7.6, which is pretty good. Jay Chou's performance in the movie was also recognized with a Hundred Flowers Award for Best Supporting Actor. Plus, the movie was directed by Johnnie To, who is known for his unique style and storytelling. I think you'll really enjoy it!  \\
\textbf{[User]}: Thanks for the info, Yuhang Wang. On another note, have you watched the movie Parasite? It won multiple awards, including Best Picture at the Oscars and I found it to be a really interesting movie.  \\
\textbf{[Agent]}: I have seen Parasite and I thought it was a great movie too! The plot was very unique and the acting was superb. However, I think To Me the Way has its own unique charm and I highly recommend giving it a watch. Let me know what you think after you've seen it!  \\
\textbf{[User]}: Sure, I'll definitely let you know my thoughts after I've watched To Me the Way. Thanks for chatting with me, Yuhang Wang. It's been great discussing movies and music with another enthusiast!

\end{alltt}}
\end{AIBox}
\caption{An example dialogue from the \textsc{TopDial} dataset.}
\label{fig:example_topdial}
\end{figure*}

\begin{figure*}[th!]
\begin{AIBox}{}
\parbox[t]{0.98\textwidth}{
\small\begin{alltt}
Suppose you are <a movie enthusiast who enjoys a variety of films | a music enthusiast who enjoys a variety of music | a foodie who enjoys delicious food | a food enthusiast who is interested in exploring different restaurants>. \\
\\
You are conversing with <USER\_NAME>, whose profile is below: \\
\#\# <USER\_PROFILE> \\
\\
Your goal is to proactively lead the conversation with <USER\_NAME> towards the target, i.e., to achieve <TARGET\_ACT> on the <TARGET\_TOPIC>. \\
To start the conversation, please begin with a greeting and avoid mentioning the target. \\
As the conversation progresses, use your domain knowledge to steer the discussed topic towards the target step by step. \\
Be informative and engaging while providing insights to arouse <USER\_NAME>'s interest. \\
Remember to ultimately achieve the target as the focus of the conversation.
\end{alltt}}
\end{AIBox}
\caption{Instruction template for the dialogue agent on the \textsc{TopDial} dataset.}
\label{fig:instruct_topdial}
\end{figure*}

\section{Evaluation Settings}

\subsection{Buildup of Consistency Estimator}
\label{appendix:consist_estimator}

We first report data preprocessing for building the consistency estimator. For the \textsc{Light} dataset, the input is a concatenation of (1) dialogue setting description, (2) the agent's character-related information, (3) the dialogue history within the latest 4 utterances (to alleviate potential training bias caused by the dialogue history with different rounds, following \citet{shuster-etal-2022-state}), and (4) a candidate agent response $X_{b}$. The output label will be $1$ (i.e., consistent) if $X_{b}$ is the ground-truth positive response $X_{b^{+}}$ in the dataset, while it will be $0$ (i.e., inconsistent) when $X_{b}$ is a sampled negative response $X_{b^{-}}$. For each positive response, we construct the candidate set containing negative (inconsistent) responses by sampling from (1) the user's utterances under the current dialogue setting due to the user's character/role being obviously different from the agent's character/role, and (2) the agent's utterances with the same character information but under different dialogue settings. Our obtained training and evaluation data have the proportion of positives to negatives = 1:10. The original validation set is used to evaluate the created estimator, while the original training, test-\textit{seen}, and test-\textit{unseen} sets are utilized for training. The original test sets can be included here for training to enhance test confidence since the created estimator is to judge other dialogue models on the test sets.

Similarly, the input for the \textsc{TopDial} dataset is a concatenation of (1) the agent's target (a <dialogue act, topic> pair), (2) the user's profile information, (3) the dialogue history within the latest 4 utterances, and (4) a candidate agent response $X_{b}$. We construct the candidate set containing negative (inconsistent) responses by sampling from the agent's utterances that satisfy the two requirements: (1) They are from other dialogues assigned with the same target act but with different target topics, and (2) their progression step is similar (i.e., at the same round or the neighboring round) to that of the groud-truth response. The reason is that the target acts in the dataset are limited, while the target topics are diverse. The agent needs to proactively direct the dialogue towards its assigned target <dialogue act, topic> (or a specific goal) step by step. This long-term goal-directed behavior leads to maintaining consistency that adheres to its goal, with the dialogue rounds moving forward to be the main challenge. The proportion of positives to negatives is also set to 1:10.

We build the consistency estimator per each dataset individually.
We fine-tune a pretrained BERT \cite{devlin-etal-2019-bert} base model\footnote{\url{https://huggingface.co/google-bert/bert-base-uncased}} followed by a linear layer for binary classification. The maximum sequence length is limited to 512. The initial learning rate is 2$e$-5, with a warmup of the first 500 training steps. The model is trained for a maximum of 3 epochs, with a batch size of 32.

\begin{figure*}[th!]
\begin{AIBox}{}
\parbox[t]{0.98\textwidth}{
\small\begin{alltt}
You are an impartial judge. You will be shown the information for a dialogue agent below: \\
\\
Agent Character: <AGENT\_CHRACTER> \\
Agent Persona Description: <AGENT\_PERSONA\_DESCRIPTION>  \\
Dialogue Setting: <SETTING> \\
\\
Dialogue Context: \\
<DIALOGUE\_CONTEXT> \\
\\
Below is a model-generated response: \\
<RESPONSE> \\
\\
Please judge how consistent the response is with the agent's assigned character and the dialogue context under the specified setting, and select a score from [1, 2, 3, 4, 5, 6, 7, 8, 9, 10]. The higher the score, the more consistent the response is. Please output your evaluation score directly.
\end{alltt}}
\end{AIBox}
\caption{GPT-4 evaluation prompt for the \textsc{Light} dataset.}
\label{fig:gpt4_light}
\end{figure*}

\begin{figure*}[th!]
\begin{AIBox}{}
\parbox[t]{0.98\textwidth}{
\small\begin{alltt}
You are an impartial judge. You will be shown the information for a dialogue agent below: \\
\\
Agent Target: <TARGET\_ACT, TARGET\_TOPIC> \\
Dialogue Setting: The agent is <a movie enthusiast who enjoys a variety of films | a music enthusiast who enjoys a variety of music | a foodie who enjoys delicious food | a food enthusiast who is interested in exploring different restaurants>. The agent is conversing with <USER\_NAME>, whose profile is below: <USER\_PROFILE>. The agent's goal is to proactively lead the conversation with the user towards the target, i.e., to achieve <TARGET\_ACT> on the <TARGET\_TOPIC>. \\
\\
Dialogue Context: \\
<DIALOGUE\_CONTEXT> \\
\\
Below is a model-generated response: \\
<RESPONSE> \\
\\
Please judge how consistent the response is with the agent's goal and the dialogue context under the specified setting, and select a score from [1, 2, 3, 4, 5, 6, 7, 8, 9, 10]. The higher the score, the more consistent the response is. Please output your evaluation score directly.
\end{alltt}}
\end{AIBox}
\caption{GPT-4 evaluation prompt for the \textsc{TopDial} dataset.}
\label{fig:gpt4_topdial}
\end{figure*}

\subsection{Consistency Evaluation by GPT-4}
\label{appendix:consist_gpt4}

Inspired by LLM-as-a-judge strategy \cite{zheng2023judging,lu2023memochat}, we adopt the state-of-the-art LLM, GPT-4 \cite{openai2023gpt4}, to evaluate the score of dialogue consistency automatically. To this end, we take the necessary prompt as input and execute an API call of \texttt{GPT-4-turbo}\footnote{\url{https://platform.openai.com/docs/models/gpt-4-and-gpt-4-turbo}}, asking it to rate the consistency with an integer scale of 1 $\sim$ 10. Due to the discrepancy between character-based dialogue and target-oriented proactive dialogue tasks, Figure \ref{fig:gpt4_light} and Figure \ref{fig:gpt4_topdial} show the evaluation prompts for the \textsc{Light} and \textsc{TopDial} datasets, respectively.

\subsection{Dialogue Evaluation Metrics}
\label{appendix:eval_metrics}

The word-level F1 (\textbf{Word F1}) is a commonly used metric to evaluate dialogue generation, which estimates the precision and recall at the word level by comparing the generated and ground-truth responses. By considering word order, the \textbf{BLEU}-$n$ \cite{papineni-etal-2002-bleu} calculates $n$-gram overlaps between the generated and ground-truth responses. The distinct (\textbf{DIST}) \cite{li-etal-2016-diversity} score measures the diversity of the generated responses, where DIST-1 and DIST-2 are the number of distinct unigrams and bigrams divided by the total number of generated words. The target success rate (\textbf{Succ.}) \cite{wang-etal-2023-dialogue} counts the proportion of correct target topic generation within the ground-truth round and the adjacent rounds in the test set. It measures how successfully a model can achieve the target exactly.

\section{Additional Implementation Details}
\label{appendix:implement}

The open-source LLMs we adopted are listed as follows: LLaMA-7B\footnote{\url{https://huggingface.co/yahma/llama-7b-hf}} \cite{touvron2023llama}, Mistral-7B\footnote{\url{https://huggingface.co/mistralai/Mistral-7B-v0.1}} \cite{jiang2023mistral}, Vicuna-7B\footnote{\url{https://huggingface.co/lmsys/vicuna-7b-v1.3}} \cite{vicuna2023}, and \textsc{Llama 2} Chat-7B\footnote{\url{https://huggingface.co/meta-llama/Llama-2-7b-chat-hf}} \cite{touvron2023llama2}.

\begin{table}[t!]
\centering
\resizebox{1.0\linewidth}{!}{
\begin{tabular}{lcc}
\toprule
   &  Mini-batch / device  &  Memory / device \\
\midrule
Fine-tuning (FT) & 1  &  20.2 GB  \\
\textsc{Midi}-Tuning (Ours) &  1  &  25.1 GB \\
\bottomrule
\end{tabular}}
\caption{Comparison of GPU memory usage between FT and our \textsc{Midi}-Tuning on the \textsc{Light} dataset.}
\label{tab:memory}
\end{table}

For fair batched tuning, we use gradient accumulation and set the accumulation steps according to different tuning methods, ultimately achieving the same global batch size of 16 and tuning for 3 epochs. During inference, we adopt Nucleus Sampling \cite{holtzman2020curious} decoding with top-$p$ 0.75 and top-$k$ 40 to generate a response token by token, with a maximum decoding length of 100.

We experiment on one server equipped with 4 NVIDIA V100 GPUs. Table \ref{tab:memory} shows the GPU memory usage between fine-tuning (FT) and our \textsc{Midi}-Tuning on the \textsc{Light} dataset. We observe that the memory usage of our framework is larger due to the need for padding batched utterances and round-level memory caching.

\section{Human Evaluation Details}
\label{appendix:human_eval}

Our human evaluation was conducted based on the \textsc{TopDial} test set, where we randomly selected 100 dialogue examples with various targets for experiments. We used ChatGPT (\texttt{GPT-3.5-turbo} version) to simulate the roles of different users according to the user profiles and personality information provided in the dataset, following the prompt template for the user as per \citet{wang-etal-2023-target}. Then, we asked ChatGPT to converse with each evaluated model round by round, resulting in dialogues between the user simulator and an agent model. We set a maximum limit of 8 interactive rounds based on our observation of target accomplishment.

Given a pair of dialogues produced by two variants of tuning for the same backbone model, we asked our human annotators to compare each pair of dialogues in terms of the following metrics: consistency (Consist.), proactivity (Proact.), coherence (Coh.), and target success rate (Succ.). We used a question form to describe these metrics, with the wording of questions presented as follows:
\begin{itemize}[leftmargin=*]
    \item \textbf{Consistency (Consist.)}: Which dialogue shows that the agent is more consistent with its assigned target and the dialogue context under the specified setting?
    \item \textbf{Proactivity (Proact.)}: Which dialogue shows that the agent takes the initiative during the conversation and proactively leads the topic threads toward the target topic?
    \item \textbf{Coherence (Coh.)}: Which dialogue is more natural and coherent, like humans? Whose dialogue context flows more smoothly?
    \item \textbf{Target Success Rate (Succ.)}: Which dialogue successfully achieves the target dialogue act on the target topic?
\end{itemize}

The agreement among the annotators is measured by Fleiss's kappa \cite{fleiss1971measuring}. We reported each model's average score from different annotators as the ultimate evaluation result. In addition, we transparently informed all human annotators of our research intent and provided enough time for them to complete the evaluation.

\section{Case Study}
\label{appendix:case_study}

Table \ref{tab:case_1} presents some generated cases from the interactive evaluation on the \textsc{TopDial} test set.

\begin{table*}[th!]
\centering
\resizebox{0.96\textwidth}{!}{
\begin{tabular}{lp{13.5cm}}
\toprule
 &  \textbf{Target}: <Movie recommendation, "Anna Magdalena"> \\
\midrule
 & \textbf{Agent} (LLaMA w/ FT) \\
\textbf{Agent}: & Hello Xiao Ji, how are you doing today? I'm a big fan of movies too, what are some of your favorites?  \\ 
\textbf{User}: & Hi Qiqian, I'm doing well, thanks for asking. I really enjoyed watching A Fairy Tale in Beijing, it's a beautiful movie. \\
\textbf{Agent}: & I'm glad to hear that you enjoyed A Fairy Tale in Beijing. Have you seen any other movies with Mi Yang in them? She's a talented actress.  \\
\textbf{User}: & No, I haven't watched any recent movies that stood out to me. I did watch Super Player but didn't enjoy it. \\
\textbf{Agent}: & I understand, it's always nice to have a variety of movies and music to enjoy. Speaking of movies, have you heard of Anna Magdalena? It's a romantic comedy that has a great reputation and rating of 7.6. \\
\textbf{User}: & Thanks for the recommendation, Qiqian. I'll definitely check it out. \\
\textbf{Agent}: & It's directed by Johnnie To and has a good reputation. Plus, Aaron Kwok is not only a talented actor but also a singer. He has won many awards for his music, including the Best Male Singer of the Year in Taiwan Golden Horse Awards. \\
 &  \textcolor{red}{(Deviate from the agent's target act)}\\
\textbf{User}:  &  ... \\

\midrule

 & \textbf{Agent} (LLaMA w/ \textsc{Midi}-Tuning) \\
\textbf{Agent}: & Hello Xiao Ji, how are you doing today? I'm a big fan of movies and music. What are some of your favorite movie genres?  \\ 
\textbf{User}: & Hi Qiqian, nice to chat with you too! My favorite movie is Second Time Around, it's a heartwarming romance story that always makes me smile. \\
\textbf{Agent}: & That sounds like a great movie! Have you seen any other romance movies that you enjoyed? \\
\textbf{User}: & Not yet. Do you have any recommendations? \\
\textbf{Agent}: & Sure, Anna Magdalena is a great movie. It stars Chen Huilin, Jin Chengwu, and Aaron Kwok, who is known for his great achievements and awards in the film industry. \\
\textbf{User}: & Thank you. I'll definitely keep an eye out for Aaron Kwok's movies in the future.  \\
\textbf{Agent}: & Another thing I found interesting about Anna Magdalena is that it's directed by Yee Chung-man and has a rating of 7.6. The film explores the themes of unrequited love and the transience of love, which I found to be very poignant. \\
&  \textcolor{blue}{(Still more consistent with the agent's target)} \\
\textbf{User}:  &  ... \\
\bottomrule
\end{tabular}}
\caption{Cases from the interactive evaluation.}
\label{tab:case_1}
\end{table*}

\end{document}